\documentclass{article}

\usepackage{multirow}

\usepackage[final, nonatbib]{neurips_2021}

\usepackage[utf8]{inputenc} 
\usepackage[T1]{fontenc}    
\usepackage{hyperref}       
\usepackage{url}            
\usepackage{booktabs}       
\usepackage{amsfonts}       
\usepackage{nicefrac}       
\usepackage{microtype}      
\usepackage{xcolor}   
\usepackage{graphicx}
\usepackage{makecell}
\usepackage{comment}
\usepackage{amsmath}
\usepackage{graphicx}
\usepackage{subfig}

\newcommand{\mat}[1]{\mathbf{#1}}

\newcommand{\lag}{\mathcal{L}}

\def\ie{{\frenchspacing\it i.e.}}
\def\eg{{\frenchspacing\it e.g.}}

\newtheorem{definition}{Definition}[section]

\title{Physics-Augmented Learning: A New Paradigm Beyond Physics-Informed Learning}

\author{%
  Ziming Liu \\
  MIT \& IAIFI \\
  zmliu@mit.edu \\
  \And
  Yunyue Chen \\
  KCL \\
  yunyue.chen@kcl.ac.uk
  \And
  Yuanqi Du \\
  GMU \\
  ydu6@gmu.edu
  \And
  Max Tegmark \\
  MIT \& IAIFI \\
  tegmark@mit.edu
}

\begin{document}

\maketitle

\begin{abstract}
   Integrating physical inductive biases into machine learning can improve model generalizability. We generalize the successful paradigm of physics-informed learning (PIL) into a more general framework that also includes what we term  physics-augmented learning (PAL). PIL and PAL complement each other by handling \textit{discriminative} and \textit{generative} properties, respectively. In numerical experiments, we show that PAL performs well on examples where PIL is inapplicable or inefficient.
\end{abstract}

\section{Introduction}
\begin{figure}[h]
    \centering
    \includegraphics[width=0.8\linewidth]{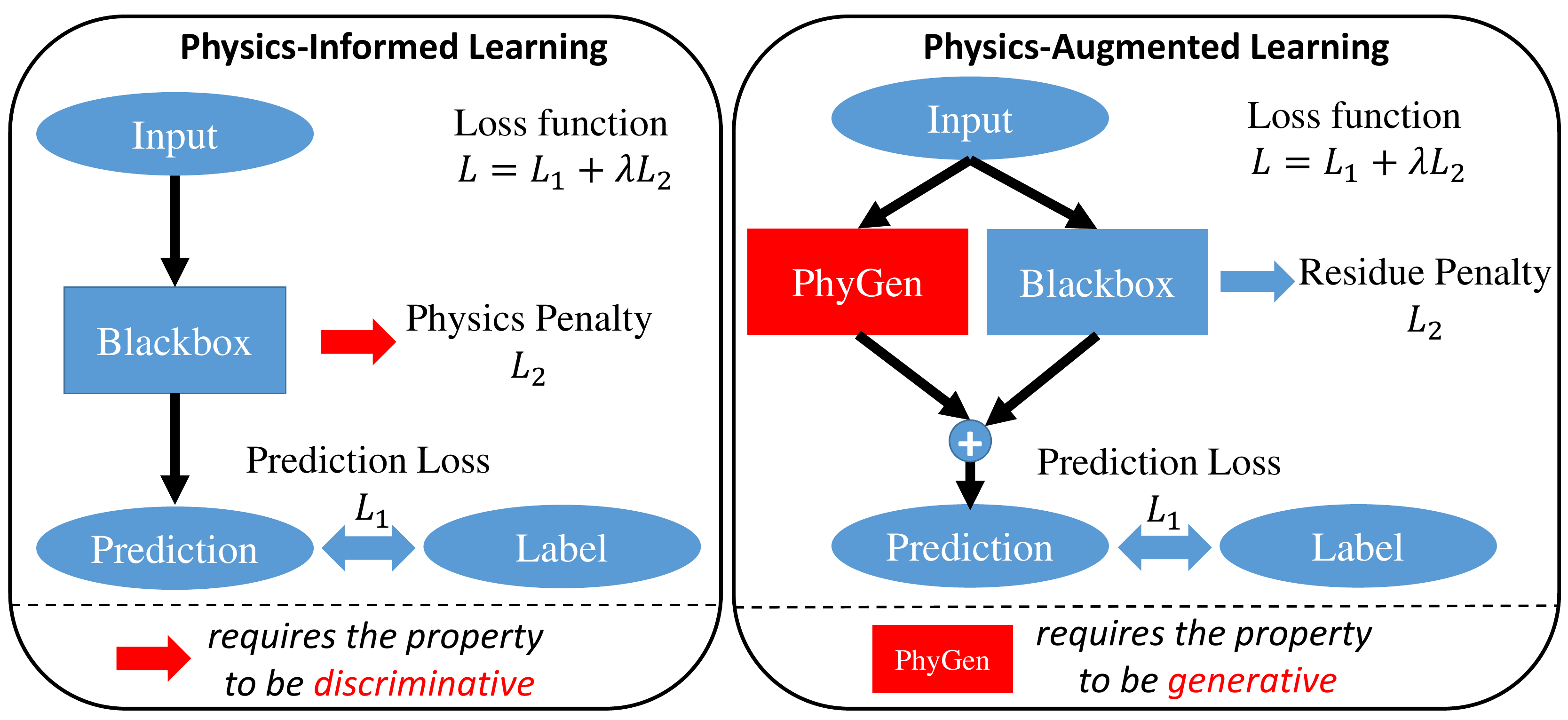}
    \caption{Compare Physics-informed learning (PIL, left) and physics-augmented learning (PAL, right). PIL and PAL apply to discriminative and generative properties respectively.}
    \label{fig:framework}
\end{figure}

Physics-informed learning (PIL)~\cite{karniadakis2021physics, raissi2017physics,raissi2017physics2,RAISSIdeep,Raissihidden} has been widely used in scientific applications where physical inductive biases are applicable. The integration of domain knowledge into machine learning can not only enhance generalization, but also make models more interpretable. 
However, PIL implicitly requires physics properties to be \textit{discriminative}, as opposed to \textit{generative} (defined below). To complement PIL, we propose a new paradigm call physics-augmented learning (PAL) to handle generative properties, as illustrated in Figure~\ref{fig:framework}. 
%
We define and compare \textit{discriminative} and \textit{generative} properties in Section~\ref{sec:dis_gen}, propose PAL in Section~\ref{sec:pil_pal} and compare it with PIL, and demonstrate PAL's effectiveness via numerical experiments in Section~\ref{sec:exp}.

\section{Discriminative and Generative Properties}\label{sec:dis_gen}

{\bf What is a property?}  A property $P$ is a mapping from an object $f$ to a boolean variable: $P(f)$ is true if $f$ satisfies $P$, false otherwise. For example, if $f$ refers to an individual with age $a$, and $P$ is the statement that ``The individual has the age no more than 30.", then $P(f)$ is true if $a\leq 30$ and $P(f)$ is false if $a>30$. 

Inspired by Generative adversarial networks (GAN)~\cite{goodfellow2014generative}, we define a \textit{generator} and a \textit{discriminator} associated with a property $P$.

\begin{definition}
{\rm \bf (Generator)}.  A generator can generate objects $f$ that have the property $P$. 
\end{definition}

\begin{definition}
{\rm \bf (Discriminator)}.  A discriminator determines if an input object $f$ has the property $P$ or not. 
\end{definition}

To be maximally useful, a generator should be implementable as a symbolic formula or feedforward neural network,
and a  discriminator can be implementable as a classifier or a loss function. More specifically, we define
\textit{ideal} generators and discriminators as follows: 

\begin{definition}
{\rm \bf (Ideal Generator)}.  An ideal generator is (1) accurate (never generate an $f$ such that $P(f)$ is false), (2) complete (can generate any $f$ such that $P(f)$ is true), (3) efficient (can generate $f$ in polynomial time), and (4) differentiable (can exploit derivative-based optimization methods such as back propagation).
\end{definition}

\begin{definition}
{\rm \bf (Ideal Discriminator)}.  An ideal discriminator is (1) accurate (always computes $P(f)$ correctly), (2) efficient (computes $P(f)$ in polynomial time), and (3) differentiable (can exploit derivative-based optimization methods such as back propagation). In this paper, we deal with one specific ideal discriminator  $\hat{L}$ such that $\hat{L}f=0$ when $P(f)$ is true and $\hat{L}f\neq 0$ when $P(f)$ is false.
\end{definition}

We now define discriminative and generative properties:

\begin{definition}
{\rm \bf (Generative property)}. A property $P$ is \textit{generative} if there exists an ideal generator for $P$.
\end{definition}

\begin{definition}
{\rm \bf (Discriminative property)}.  A property $P$ is \textit{discriminative} if there exists an ideal discriminator for $P$.
\end{definition}

Let us clarify these abstract definitions with a few examples of properties below, summarized in Table \ref{tab:properties}).

{\bf A. Lagrangian property}
For a physics system with generalized coordinate $\mat{q}$ and velocity $\dot{\mat{q}}$, the acceleration field $\ddot{\mat{q}}$ is {\it Lagrangian} if there exists a Lagrangian function $\lag(\mat{q},\dot{\mat{q}})$ such that
$\ddot{\mat{q}}=(\nabla_{\dot{\mat{q}}}\nabla_{\dot{\mat{q}}}^T \lag)^{-1}(\nabla_{\mat{q}}\lag-(\nabla_{\mat{q}}\nabla^T_{\dot{\mat{q}}}\lag)\dot{\mat{q}})$ ~\cite{cranmer2020lagrangian, nnphd}. The Lagrangian property is generative by definition but not discriminative, because (perhaps surprisingly) there is no known efficient method to determine whether a given $\ddot{\mat{q}}$ is Lagrangian or not~\cite{nnphd}.

{\bf B. Positive definiteness} By analogy with linear operators, we say that a function $f(x)$ is {\it positive definite} if there exists a function $g$ such that $f(x)=g(g(x))$. For example, the function that time-evolves a physical system during an interval $\Delta T$ is positive definite, since it is equivalent by time-evolving by $\Delta T/2$ twice.
The positive definiteness  property is generative by definition, but not discriminative to the best of our knowledge. 

{\bf C. Manifest symmetry} Many manifest symmetries are discriminative, with associated discriminators that can be elegantly described by partial differential equations~\cite{liu2021machinelearning}. For example,
a vector field $\mat{f}(\mat{x})$ is symmetric under a Lie group $\mathcal{G}$ if
 $\mat{f}(g\mat{x})=g(\mat{f}(\mat{x}))$ for all $g\in\mathcal{G}$.
 This symmetry property is discriminative because it is equivalent to 
 $(K_i\mat{x}\cdot\nabla-K_i)\mat{f}=0$ for all group generators $K_i$\cite{liu2021machinelearning}. 
 It is also generative due to recent advances in equivariant neural networks~\cite{cohen2016group, thomas2018tensor,fuchs2020se,kondor2018clebsch,satorras2021n}. 

{\bf D. Hidden symmetry} However, hidden symmetries are not discriminative, because they require coordinate transformations to `generate' manifestly symmetric objects~\cite{liu2021machinelearning}. For example, manifest Hamiltonicity is shown to be discriminative in~\cite{liu2021machinelearning}, but hidden Hamiltonicity is a generative property and that is equivalent to Lagrangianess \cite{nnphd}.

{\bf E. Separability}: A differentiable bivariate function $f(x_1,x_2)$ is (additively) separable if there exists two unary functions $f_1$ and $f_2$ such that $f(x_1,x_2)=f_1(x_1)+f_2(x_2)$. Separability is generative by definition, and also discriminative because it is equivalent to  $\hat{L}f\equiv\partial^2 f/\partial x_1\partial x_2=0$ ~\cite{Udrescueaay2631, udrescu2020ai}.

{\bf F. PDE satisfiability} We say a function $f(x,t)$ satisfies a partial differential equation (PDE) if $g(t,f,f_t,f_x,..)=0$. This property is discriminative by definition, with $\hat{L}f=g$. It is also generative when $f(x,t)$ can be efficiently computed by a numerical PDE solver given proper boundary conditions; this idea underlies neural ordinary/partial/stochastic differential equations~\cite{chen2019neural,hsieh2019learning,kidger2021neural}.

\begin{table}[]
    \centering
    \caption{Generative and discriminative properties}
    \resizebox{\textwidth}{!}{%
    \begin{tabular}{|c|c|c|c|c|c|c|c|}\hline
    Properties     &  \makecell{A. Lagrangian \\ Property} & \makecell{B. Positive \\ Definiteness}  &  \makecell{C. Manifest \\ Symmetry} &  \makecell{D. Hidden \\ Symmetry}  &   E. Separability & \makecell{F. PDE\\ Satisfiability} \\\hline
    Generative & Yes & Yes & Yes & Yes  & Yes & Yes \\\hline
    Discriminative & No & No & Yes & No & Yes & Yes \\\hline
    \end{tabular}}
    \label{tab:properties}
\end{table}

\section{Physics-informed Learning (PIL) and Physics-augmented learning (PAL)}\label{sec:pil_pal}

In this section, we first review physics-informed learning (PIL) and its limitations, motivating our proposed physics-augmented learning (PAL) framework.

\subsection{Physics-informed Learning (PIL)}

The essence of PIL is to seamlessly integrate data and mathematical
physics models, and a common way is to add a soft penalty term $L_2$ (corresponding to physics properties) to the  prediction loss $L_1$~\cite{karniadakis2021physics} (Figure \ref{fig:framework}, left panel). PIL works for {\it discriminative} properties. Indeed, one of its greatest successes lies in solving forward/inverse PDE problems~\cite{raissi2017physics,raissi2017physics2,Raissihidden}, based on the unstated fact that satisfying a PDE is discriminative. We clarify PIL with a toy example below.

{\bf Example: PIL for separability} Suppose that our training dataset ($N$ samples) is generated by the oracle $y=f_0(x_1,x_2)$ where $f_0:\mathbb{R}^2\to\mathbb{R}$ and that we want to use a parametrized neural network $f(x_1,x_2;\theta)$ to fit the data with a function $f(x_1,x_2;\theta)$ that is additively separable. 
PIL does this using a loss function with two terms: $L\equiv L_1+\lambda L_2$, 
where the prediction loss $L_1$ and separability loss $L_2$ are 
$$
L_1(\theta)\equiv\frac{1}{N}\sum_i |f_0(x_1^{(i)},x_2^{(i)})-f(x_1^{(i)},x_2^{(i)};\theta)|,
\quad
L_2(\theta)\equiv\frac{1}{N}\sum_i\left| {\partial^2 f(x_1^{(i)},x_2^{(i)};\theta)\over\partial x_1\partial x_2}\right|,
$$
and the constant $\lambda>0$ is a penalty coefficient.\footnote{Alternatively, $L_2$ can be expressed in terms of finite differences instead of derivatives; we do this for our numerical experiments. For example, additive separability corresponds to the condition listed in Table 2, which we average over all pairs of data points.} 


By definition, each discriminative property $P$ can be written $\hat{L}f=0$ for some operator $\hat{L}$,
so $L_2\equiv |\hat{L}f|$ is a natural measure of property violation. 
In contrast, non-discriminative properties, \eg,  being Lagrangian or positive definite, lack a known efficiently  computable criterion $\hat{L}f=0$.
Fortunately, the PAL framework proposed below can come to rescue whenever the properties of interest are {\it generative}.

\begin{table}[]
    \centering
    \caption{Neural network and loss function for PIL and PAL on the separability example}
    \resizebox{\textwidth}{!}{%
    \begin{tabular}{|c|c|c|}\hline
    Paradigm & NN & \makecell{Prediction error $L_1$, \\Penalty $L_2$ }
     \\\hline
    PIL & $f(x_1,x_2;\theta)$ & \makecell{$\frac{1}{N}\sum_{i}|f_0(x_1^{(i)},x_2^{(i)})-f(x_1^{(i)},x_2^{(i)};\theta)|$, \\ $\frac{2}{N(N-1)}\sum_{j>i}|f(x_1^{(i)},x_2^{(i)};\theta)+f(x_1^{(j)},x_2^{(j)};\theta)-f(x_1^{(i)},x_2^{(j)};\theta)-f(x_1^{(j)},x_2^{(i)};\theta)|$} \\\hline
    PAL & \makecell{$f_1(x_1;\theta_1),f_2(x_2;\theta_2),$ \\ $f_{12}(x_1,x_2;\theta_{12})$} & \makecell{$\frac{1}{N}\sum_i|f_0(x_1^{(i)},x_2^{(i)})-(f_1(x_1^{(i)};\theta_1)+f_2(x_2^{(i)};\theta_2))-f_{12}(x_1^{(i)},x_2^{(i)};\theta_{12})|$, \\ $\frac{1}{N}\sum_i|f_{12}(x_1^{(i)},x_2^{(i)};\theta_{12})|$}  \\\hline
    \end{tabular} }
    \label{tab:separability}
\end{table}

\subsection{Physics-augmented Learning (PAL)}

Although both PIL and PAL aim to leverage inductive biases in machine learning, their approaches are quite different: while PIL is based on regularization design, PAL is based on model design. Useful regularization designs and
model designs are only available for properties that are discriminative  and generative, respectively.

In the PAL paradigm, the whole model consists of two parallel modules (see Figure \ref{fig:framework}, right panel): the first module ({\tt PhyGen}) strictly satisfies the generative property, and the second module ({\tt Blackbox}) \textit{augments} the expressive power to allow violation of the property. The loss function consists of two terms: the standard prediction error $L_1$, and a penalty term $L_2$ defined as some norm of {\tt Blackbox} module output. The combined loss function is $L=L_1+\lambda L_2$.

{\bf Example: PAL for separability} In PAL we have two neural modules {\tt PhyGen} and {\tt Blackbox}. {\tt PhyGen} satisfies strictly the additive separability by having two sub-networks $(f_1(x_1;\theta_1), f_2(x_2;\theta_2))$ and outputs their sum. {\tt Blackbox} is a fully-connected neural network $f_{12}(x_1,x_2;\theta_{12})$ that can universally approximate any two-variable continuous function. The whole prediction is thus $(f_1(x_1;\theta_1)+f_2(x_2;\theta_2))+f_{12}(x_1,x_2;\theta_{12})$ and the prediction loss $L_1$ is its distance from the label $y=f_0(x_1,x_2)$. The penalty loss is simply the function norm of the {\tt Blackbox} \ie, $L_2=|f_{12}(x_1,x_2;\theta_{12})|$. 
Our PIL and PAL examples are summarized and compared in Table \ref{tab:separability}.

{\bf When should I use PAL rather than PIL?} PIL and PAL are complementary frameworks exploiting discriminative and generative properties, respectively. As a consequence, one should resort to PAL without hesitation 
when the property of interest is generative and non-discriminative. Another reason to use PAL is that it provides a model decomposition into {\tt PhyGen} and {\tt Blackbox}, and interpreting them potentially produces physical insights. For example, the decomposition into conservative and non-conservative force fields enabled new-physics discovery in~\cite{nnphd}.

{\bf How to choose the hyperparameter $\lambda$?} It was recently proven that 
$L=L_1+\lambda L_2$ produces a phase transition at $\lambda=1$ ~\cite{nnphd}\footnote{Here the loss functions $L_1$ and $L_2$ should be defined as norms to produce sharp phase transition behavior, \eg, mean-absolute error (MAE) or Euclidean norm. 
In contrast, MSE does not produce a sharp phase transition.}. $\lambda>1$ is the undesirable phase, so in principle one can simply choose any $\lambda<1$ to obtain vanishing prediction loss; the numerical results of  \cite{nnphd} suggest that $\lambda\in[0.02,0.5]$ produces accurate and robust results in practice. We verify these observations with the additive separability example in Appendix~\ref{app:symbolic}. In our numerical experiments, we choose $\lambda=0.2$ if not stated otherwise.

\section{Numerical Experiments}\label{sec:exp}

We demonstrate the effectiveness of PAL on two tasks: symbolic regression and dynamics prediction. PAL performs well on these applications, while PIL is inapplicable or performs worse than PAL.

\subsection{Symbolic Regression}

The goal of symbolic regression is to find a symbolic expression that matches data from an unknown function $f$. The physics-inspired AI Feynman  symbolic regression module~\cite{Udrescueaay2631,udrescu2020ai}
 tests if a dataset satisfies certain properties, including symmetries. However, AI Feynman can only 
discovers and exploits these if they hold with high accuracy. 
To relax this, we employ PAL to first decompose the function $f$ into two parts: a property-satisfying part $f_+$ and a property-violating residual $f_-$. We then apply AI Feynman to both parts separately to obtain symbolic formulas; $f_+$ satisfies the strict property which AI Feynman can exploit.

We experiment with three properties: additive separability, rotational invariance and positive definiteness. Training details can be found in Appendix \ref{app:symbolic}. Table \ref{tab:symbolic} shows the symbolic regression results: for the first two properties, PAL decomposes the function as desired while PIL can only learn the whole function; for the positivity example, PIL is not applicable while PAL can extract meaningful partial function $g$.

\begin{table}[]
    \centering
    \caption{Symbolic regression results}
    \begin{tabular}{|c|c|c|c|}\hline
    Property & Methods & $f_+$ & $f_-$  \\\hline
    \multirow{3}{*}{\makecell{Additive Separability \\ $f=x_1^2+x_2^2+x_1x_2$}}  & Truth & $(x_1^2)+(x_2^2)$ & $x_1x_2$ \\\cline{2-4}
    & AI Feynman + PAL & $(x_1^2-0.02)+ (x_2^2-0.01)$ & $x_1x_2+0.03$
    \\\cline{2-4}
    & AI Feynman + PIL & \multicolumn{2}{c|}{$x_1^2+x_2^2+x_1x_2$} \\\hline
    \multirow{3}{*}{\makecell{Rotational Invariance \\ $f=0.5(x_1^2+x_2^2)+0.32x_1$}}  & Truth & $0.5(x_1^2+x_2^2)$ & $0.32x_1$ \\\cline{2-4}
    & AI Feynman + PAL & $0.5(x_1^2+x_2^2)$ & $0.31998x_1$
    \\\cline{2-4}
    & AI Feynman + PIL & \multicolumn{2}{c|}{$0.5(x_1^2+x_2^2)+0.32x_1$} \\\hline
    \multirow{3}{*}{\makecell{Positivity \\ $f={\rm sin}({\rm sin}(x))+0$}} & Truth & ${\rm sin}({\rm sin}(x))$ & $0$ \\\cline{2-4}
    & AI Feynman + PAL & $g(g(x)), g=-{\rm sin}(x)+0.004$ & $0$ \\\cline{2-4}
    & AI Feynman + PIL & \multicolumn{2}{c|}{Not Applicable} \\\hline
    \end{tabular}
    \label{tab:symbolic}
\end{table}

\begin{figure}[htbp]
    \vskip -0.7cm
	\centering
    \begin{minipage}{0.5\textwidth}
        \subfloat[Trajectory]
        {
            \captionsetup[subfigure]{labelformat=empty}
            \subfloat[Ground Truth]{
                \includegraphics[width=0.48\linewidth]{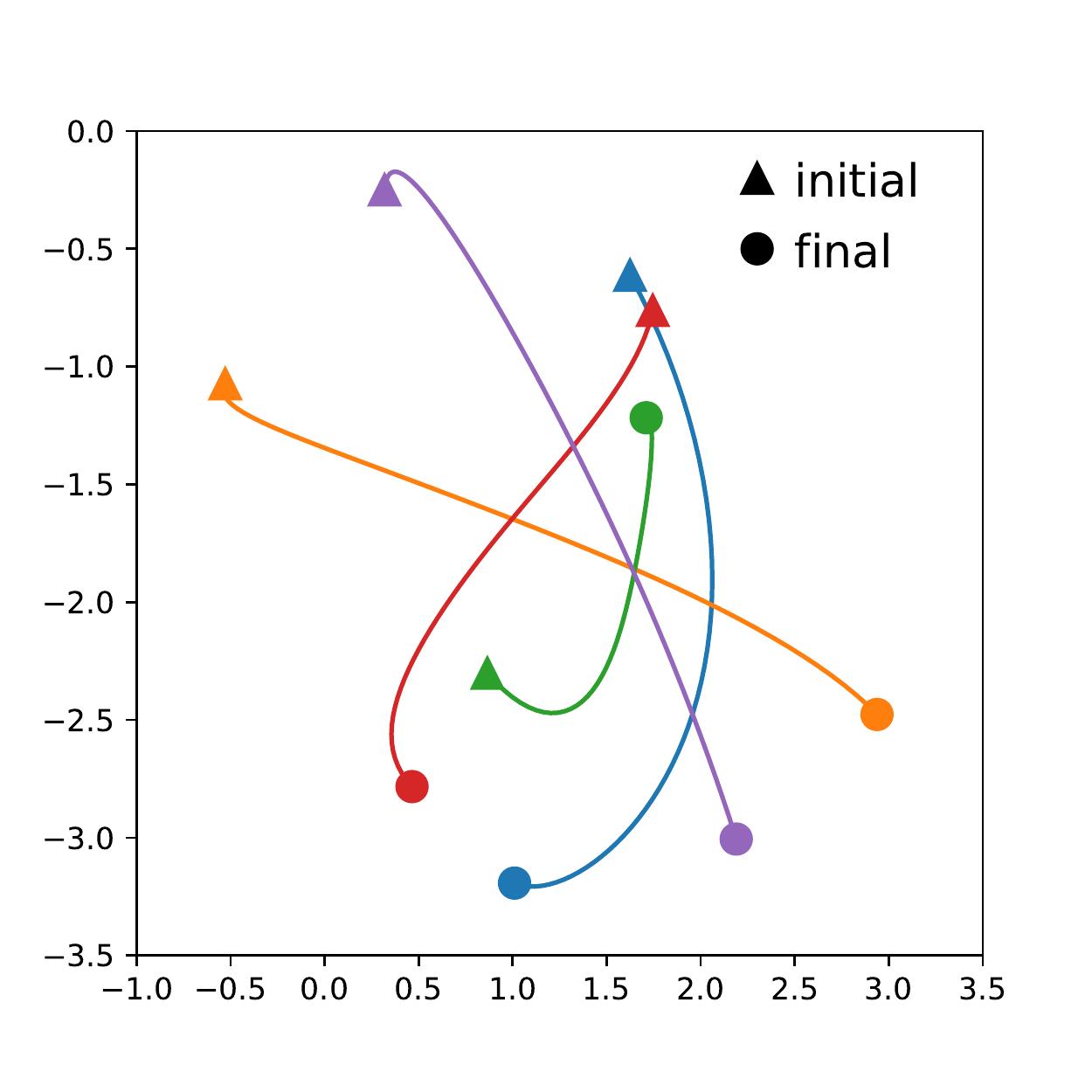}
            }
            \subfloat[PAL \& PIL]{
                \includegraphics[width=0.48\linewidth]{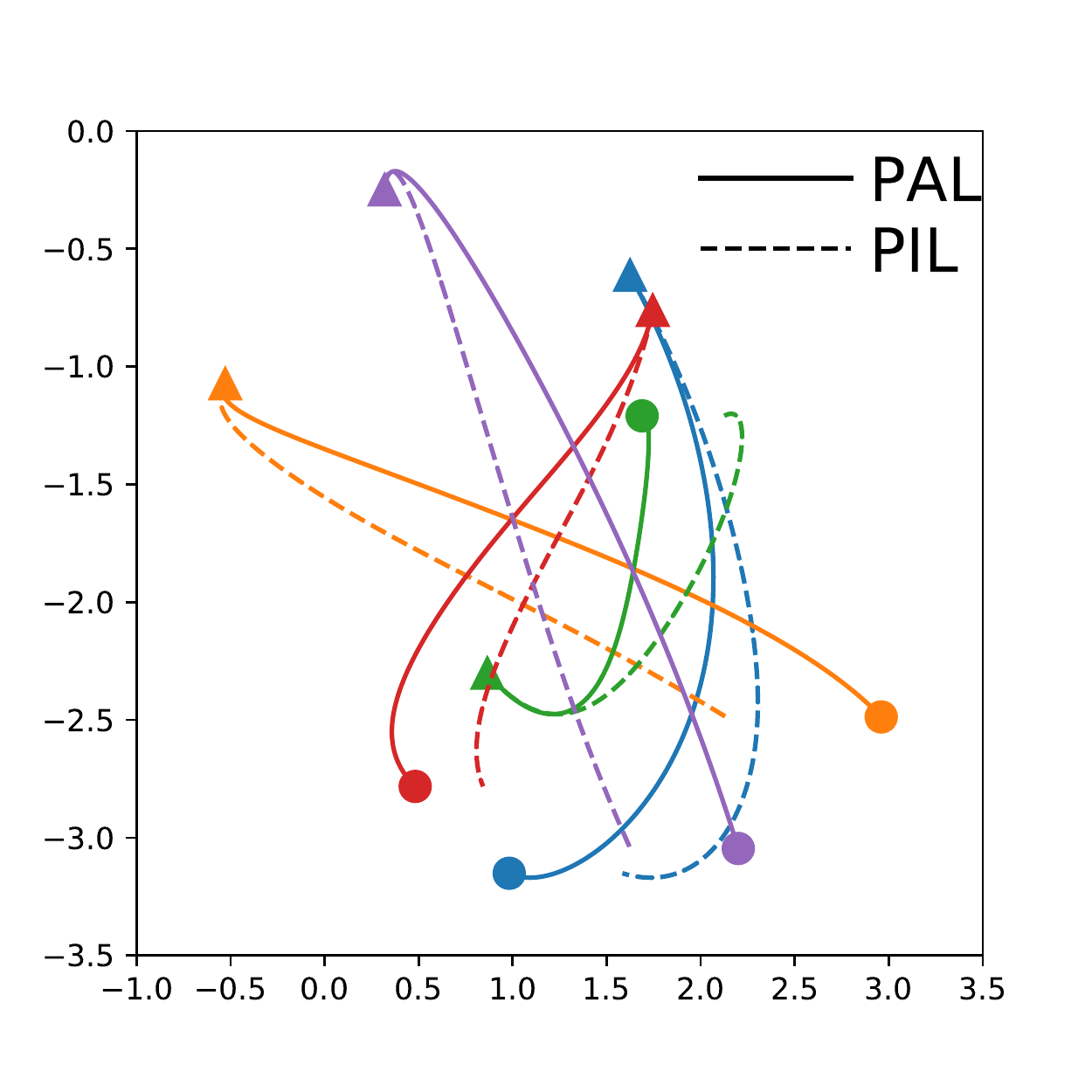}
            }
            \setcounter{subfigure}{1}
            \label{fig2:a}
        }
    \label{Fig2_a}
    \end{minipage}
    \begin{minipage}{0.35\textwidth}
        \center
        \subfloat
        {
            \captionsetup[subfigure]{labelformat=empty}
            \subfloat[(b) Force]{
                \includegraphics[width=0.96\linewidth]{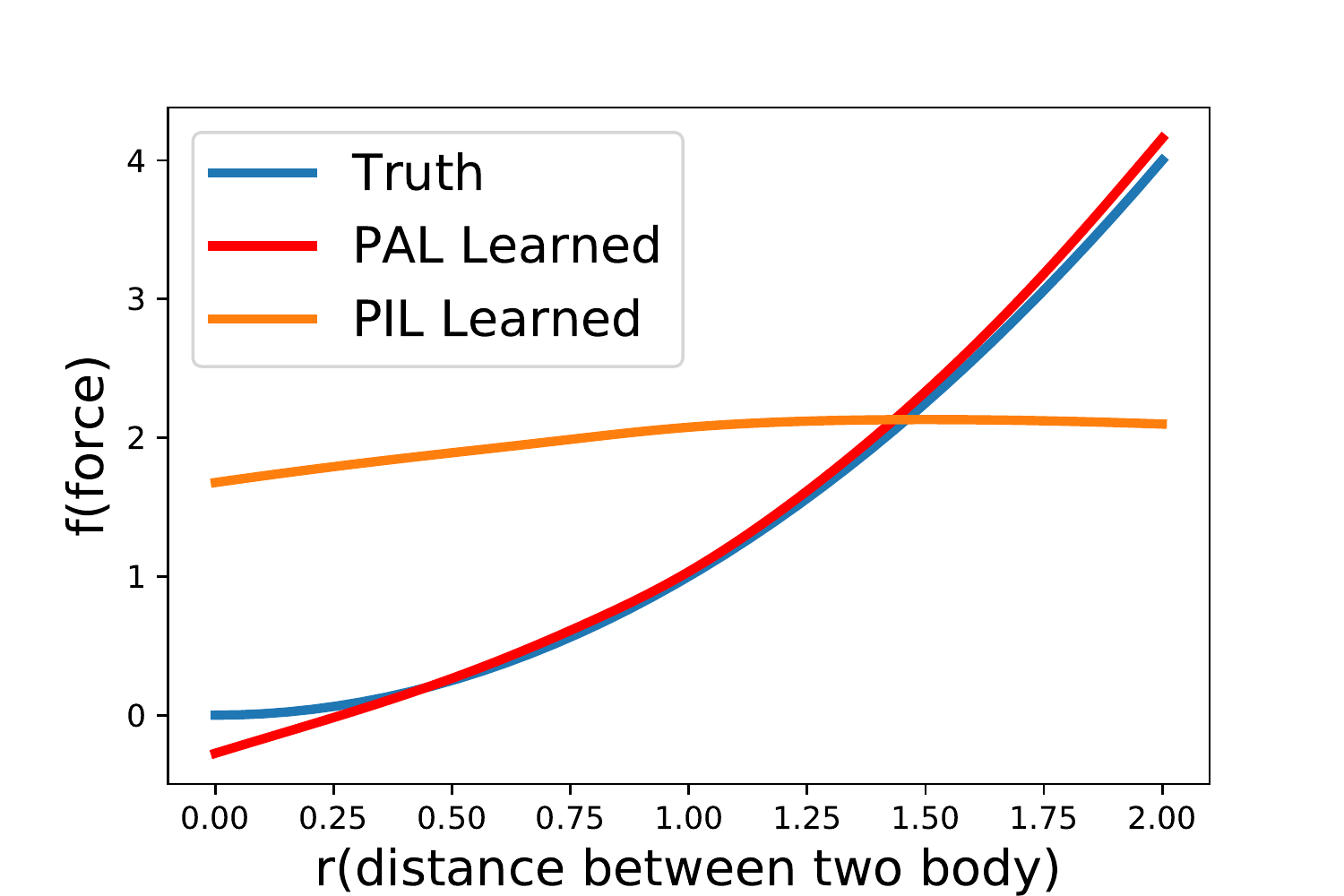}
            }
        
        }
    \end{minipage}
    \caption{Dynamics prediction results: (a) Trajectory; (b) Force.}
    \label{fig2}
    \vskip -0.5cm
\end{figure}

\subsection{Dynamics prediction: N-body Problem}

In this example, we aim to learn $N$-body dynamics from the initial $(t=0)$ and final $(t=1)$ states of $N$ particles. We assume that these unit mass particles ($m=1$) obey the Newtonian mechanics with pairwise interactions, \ie, particle $i$ exerts a force $\mat{f}_{ij}=f(|\mat{x}_i-\mat{x}_j|)\frac{\mat{x}_i-\mat{x}_j}{|\mat{x}_i-\mat{x}_j|_2}$ on particle $j$. Our ground truth is that $f(r)=r^2$ when computing final states from initial states, but we pretend not to know $f$ and aim to infer it solely from the initial and final states (indicated by triangles and dots in Figure 2).
The property $P$ that we explore with PAL and PIL is the time-independence of $f$, which is clearly both discriminative and generative.  Details of loss functions, data generation and training details are included in Appendix \ref{app:nn+losses} and  \ref{app:n-body}.
Figure 2 shows that PAL outperforms PIL in terms of both trajectory interpolation and force recovery. 
In other words, although both PIL and PAL can be applied for this example, PAL reveals the underlying dynamics much more accurately that PIL.

\section{Conclusions \& Discussions}

We have proposed a new paradigm called physics-augmented learning (PAL) to effectively integrate physical properties into unconstrained neural networks. PAL complements the well-known physics-informed learning (PIL) paradigm, by applying in some cases where PIL is inapplicable and by outperforming PIL in some cases where both can be used. While PIL is based on regularization design and applies to discriminative properties, PAL is based on model design and applies to generative properties. 

Although PAL in its general form is explicitly formulated for the first time in this paper, examples of it have been implicitly adopted in many successful machine learning models owing to its ability to integrate human knowledge into the model design phase. For example, AlphaFold2 demonstrated that designing deep learning models with proper inductive biases could give superior performance for an unsolved grand challenge \cite{jumper2021highly}. There is growing interest in the ML community in how inductive biases can shape deep learning models; for example, \cite{raghu2021vision} discovers how two fundamental deep learning models (CNNs and Transformers) leverage their own inductive biases and tackle challenges in the computer vision domain. 
We are hopeful that the PAL-PIL framework unifying inductive biases can
lead to further advances in the ML community for tackling real-world challenges with optimal selection and design of inductive biases.


\section*{Acknowledgement}
We would like to thank Silviu Udrescu and Jiaxi Zhao for valuable discussions. We thank the Center for Brains, Minds, and Machines (CBMM)
for hospitality. This work was supported by The Casey
and Family Foundation, the Foundational Questions Institute, the Rothberg Family Fund for Cognitive Science and IAIFI through NSF grant PHY-2019786.


\bibliography{main}
\bibliographystyle{unsrt}

\clearpage

\begin{center}
    {\LARGE\bf  Supplementary material}
\end{center}

\begin{appendix}

\section{Definitions of PIL and PAL losses}\label{app:nn+losses}

In Section \ref{sec:exp}, we conducted experiments related to four properties: additive separability, rotational invariance, positivity and time independence. PAL applies to all of these properties, while PIL is not applicable to positivity but applies to other three properties. In Table \ref{tab:nn+losses} we show the design of neural networks and loss functions for each property in the framework of PIL and PAL. 

All neural networks $f_{*}(*;*)$ (each $*$ can be anything) are fully-connected networks which can approximate any continuous function of input variables (if wide enough). Their differences lie in input variables. For PIL, the only network is fully-connected and takes in all available variables. For PAL, we have a {\tt PhyGen} and {\tt Blackbox} which usually have different inputs. {\tt Blackbox} takes in all available input variables while {\tt PhyGen} usually takes in fewer variables which is determined by the specific property, as detailed below.

\begin{itemize}
    \item Additive separability. {\tt PhyGen}: $f_1$ takes in $x_1$ only, while $f_2$ takes in $x_2$ only. The output is $f_1(x_1;\theta_1)+f_2(x_2;\theta_2)$. {\tt Blackbox}: $f_{12}$ takes in both $x_1$ and $x_2$. The output is $f_{12}(x_1,x_2;\theta_{12})$.
    \item Rotational invariance. {\tt PhyGen}: $f_1$ takes in one variable $R\equiv\sqrt{x_1^2+x_2^2}$ only. The output is $f_1(R;\theta_1)$. {\tt Blackbox}: $f_2$ takes in both $x_1$ and $x_2$. The output is $f_2(x_1,x_2;\theta_2)$.
    \item Positivity. {\tt PhyGen}: $f_1$ takes in $x$ and outputs $f_1(f_1(x;\theta_1);\theta_1)$. {\tt Blackbox}: $f_2$ takes in $x$ and outputs $f_2(x;\theta_2)$. 
    \item Time independence. {\tt PhyGen}: $f_1$ takes in $r$ where $r$ is the distance between two particles. The output is $f_1(r;\theta_1)$. {\tt Blackbox}: $f_2$ takes in both $r$ and $t$. The output is $f_2(r,t;\theta_2)$. 
\end{itemize}

\begin{table}[h]
    \centering
    \caption{PIL and PAL neural networks and loss functions for four properties}
    \resizebox{\textwidth}{!}{
    \begin{tabular}{|c|c|c|c|c|}\hline
    Property & Variables & Paradigm & NN & \makecell{Prediction error $L_1$, \\Penalty $L_2$ }
    \\\hline
    \multirow{3}{*}{\makecell{Additive \\ Separability}} & \multirow{3}{*}{$x_1,x_2$} & PIL & $f(x_1,x_2;\theta)$ & \makecell{$\frac{1}{N}\sum_i|f_0(x_1,x_2)-f(x_1,x_2;\theta)|$, \\ $\frac{2}{N(N-1)}\sum_{j>i}|f(x_1^{(i)},x_2^{(i)};\theta)+f(x_1^{(j)},x_2^{(j)};\theta)-f(x_1^{(i)},x_2^{(j)};\theta)-f(x_1^{(j)},x_2^{(i)};\theta)|$} \\\cline{3-5}
     & & PAL & \makecell{$f_1(x_1;\theta_1),f_2(x_2;\theta_2)$, \\ $f_{12}(x_1,x_2;\theta_{12})$} & \makecell{$|f_0(x_1,x_2)-(f_1(x_1;\theta_1)+f_2(x_2;\theta_2))-f_{12}(x_1,x_2;\theta_{12})|$, \\ $|f_{12}(x_1,x_2;\theta_{12})|$}  \\\hline
    \multirow{3}{*}{\makecell{Rotational\\ Invariance}} & \multirow{3}{*}{$x_1,x_2$} & PIL & \makecell{$f(x_1,x_2;\theta)$} & \makecell{$|f_0(x_1,x_2)-f(x_1,x_2;\theta)|$, \\ 
    $|f(x_1,x_2;\theta)-f(x_1',x_2';\theta)|$ [$(x_1',x_2')^T={\rm Rotation}(\alpha)(x_1,x_2)^T$]} \\\cline{3-5} & & PAL & \makecell{$f_1(R;\theta_1)$, \\  $f_2(x_1,x_2;\theta_2)$} & \makecell{$|f_0(x_1,x_2)-f_1(R;\theta_1)-f_2(x_1,x_2;\theta_2)|$, \\ $|f_2(x_1,x_2;\theta)|$} \\\hline
    \multirow{3}{*}{Positivity} & \multirow{3}{*}{$x$} & PIL & \makecell{Not \\ Applicable} & \makecell{Not \\ Applicable} \\\cline{3-5} & & PAL & \makecell{ $f_1(x;\theta_1)$,\\ $f_2(x;\theta_2)$}
    & \makecell{$|f_0(x)-f_1(f_1(x;\theta_1);\theta_1)-f_2(x;\theta_2)|$, \\ $|f_2(x;\theta_2)|$}  \\\hline
    \multirow{3}{*}{\makecell{Time \\ Independence}} & \multirow{3}{*}{$r,t$} & PIL & $f(r,t;\theta)$ & \makecell{$|\mat{z}(t=0)+\int_{0}^1 dt \mat{F}(\mat{z},t)-\mat{z}(t=1)|$,\\ $\sum_{i=1}^{n}|f(r,i\Delta t;\theta)-f(r,(i+1)\Delta ;\theta)|$} \\\cline{3-5}
    & & PAL & \makecell{$f_1(r;\theta_1)$,\\ $f_2(r,t;\theta_2)$} & \makecell{$|\mat{z}(t=0)+\int_{0}^1 dt (\mat{F}_1(\mat{z})+\mat{F}_2(\mat{z},t))-\mat{z}(t=1)|$, \\ $|f_{2}(\mat{z},t;\theta_{2})|$}  \\\hline
    \end{tabular}}
    \vskip 0.5cm
    \label{tab:nn+losses}
\end{table}

\section{Symbolic Regression Details}\label{app:symbolic}

{\bf Neural network architecture}: In the symbolic regression experiment, all the neural networks are fully-connected networks, which have 1 or 2 input neurons (depends on the number of input variables), 2 hidden layers ($\text{width}=256$) with LeakyReLU $(\alpha=0.2)$, and a single output neuron.

{\bf Additive separability}: The function $f_0(x_1,x_2)=(x_1^2+x_2^2)+x_1x_2$ can be decomposed into the additively separable part $x_1^2+x_2^2$ and a violation part $x_1x_2$. 

\begin{figure}[htbp]
	\centering
    \begin{minipage}{0.58\textwidth}
        \subfloat[Ground Truth]
        {
            \captionsetup[subfigure]{labelformat=empty}
            \subfloat[$f_1=x_1^2$]{
                \includegraphics[width=0.31\linewidth]{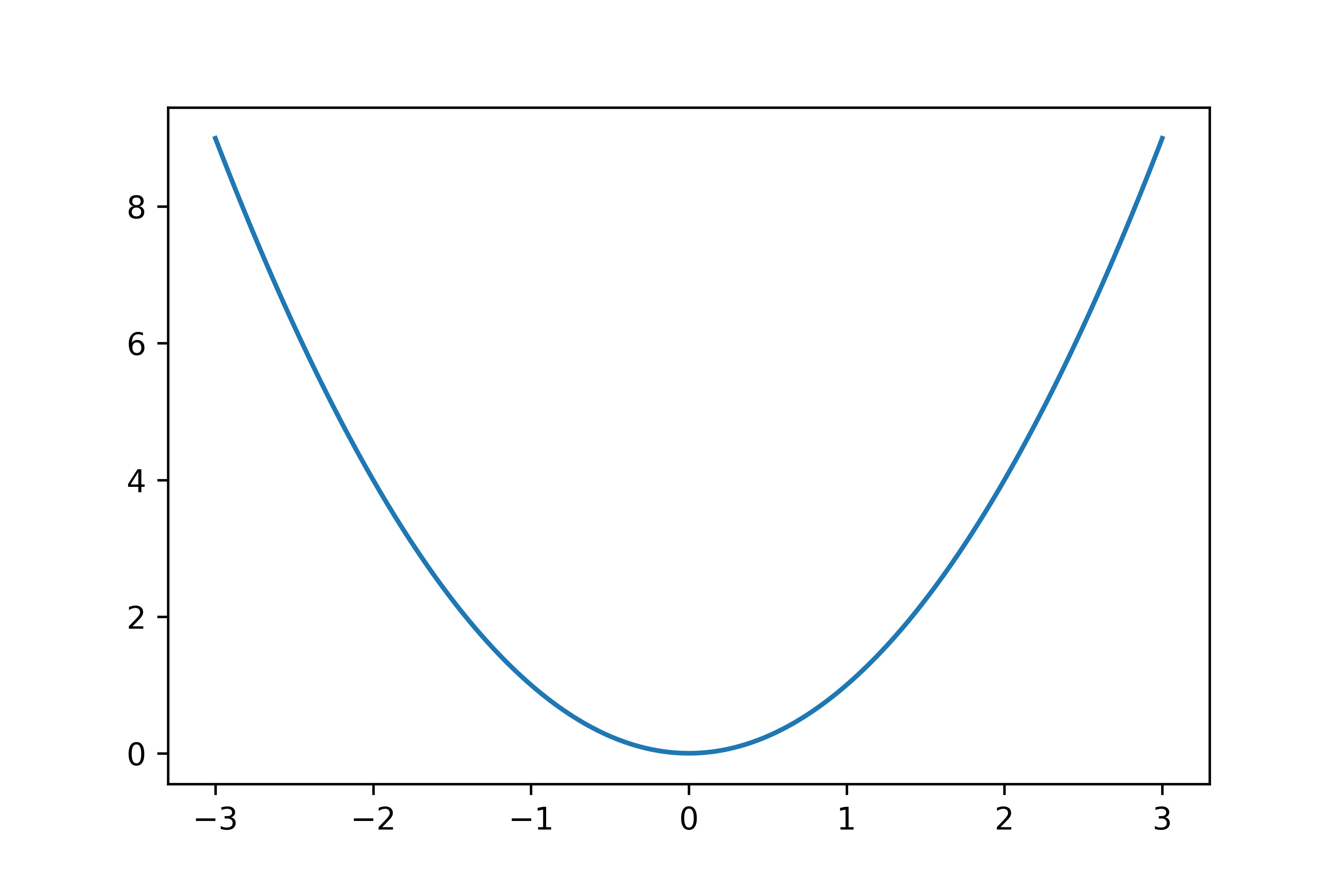}
            }
            \subfloat[$f_2=x_2^2$]{
                \includegraphics[width=0.31\linewidth]{./y_x_2.png}
            }
            \subfloat[$f_3=x_1x_2$]{
                \includegraphics[width=0.31\linewidth]{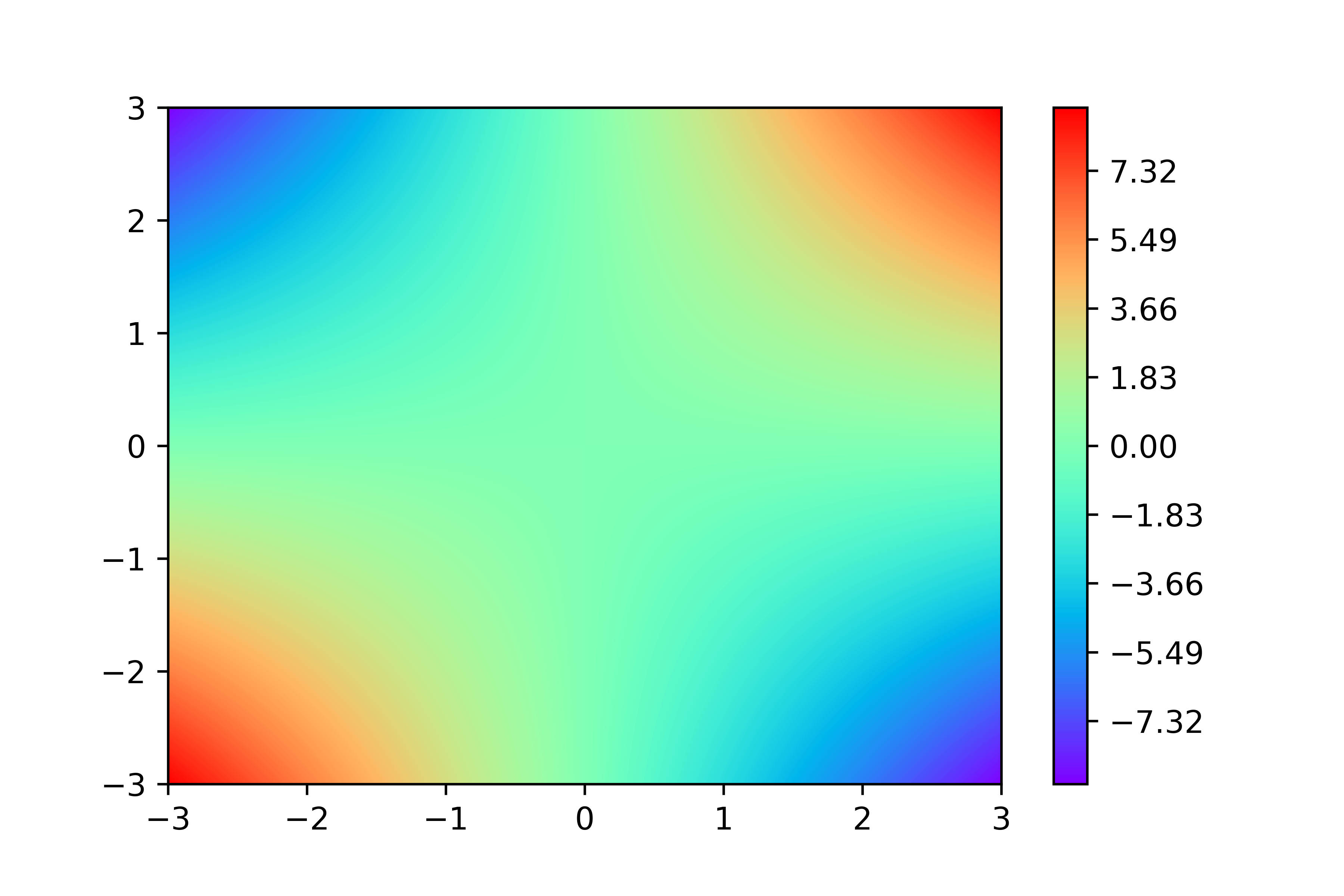}
            }
            \setcounter{subfigure}{1}
            \label{fig3:a}
        }\\
        \subfloat[Neural Networks Learned]
        {
            \captionsetup[subfigure]{labelformat=empty}
            \subfloat[$f_1(x_1;\theta_1)$]{
                \includegraphics[width=0.31\linewidth]{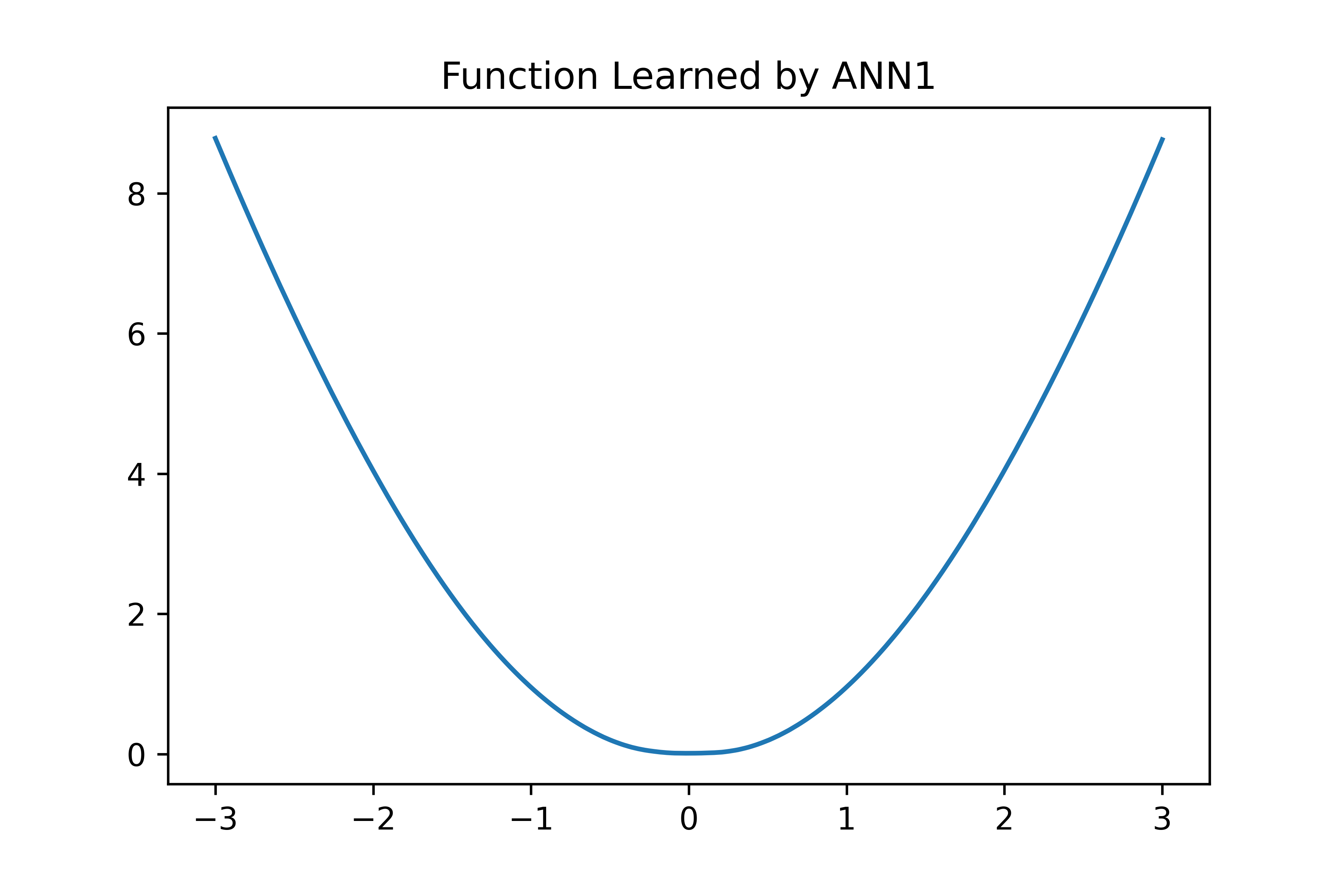}
            }
            \subfloat[$f_2(x_2;\theta_2)$]{
                \includegraphics[width=0.31\linewidth]{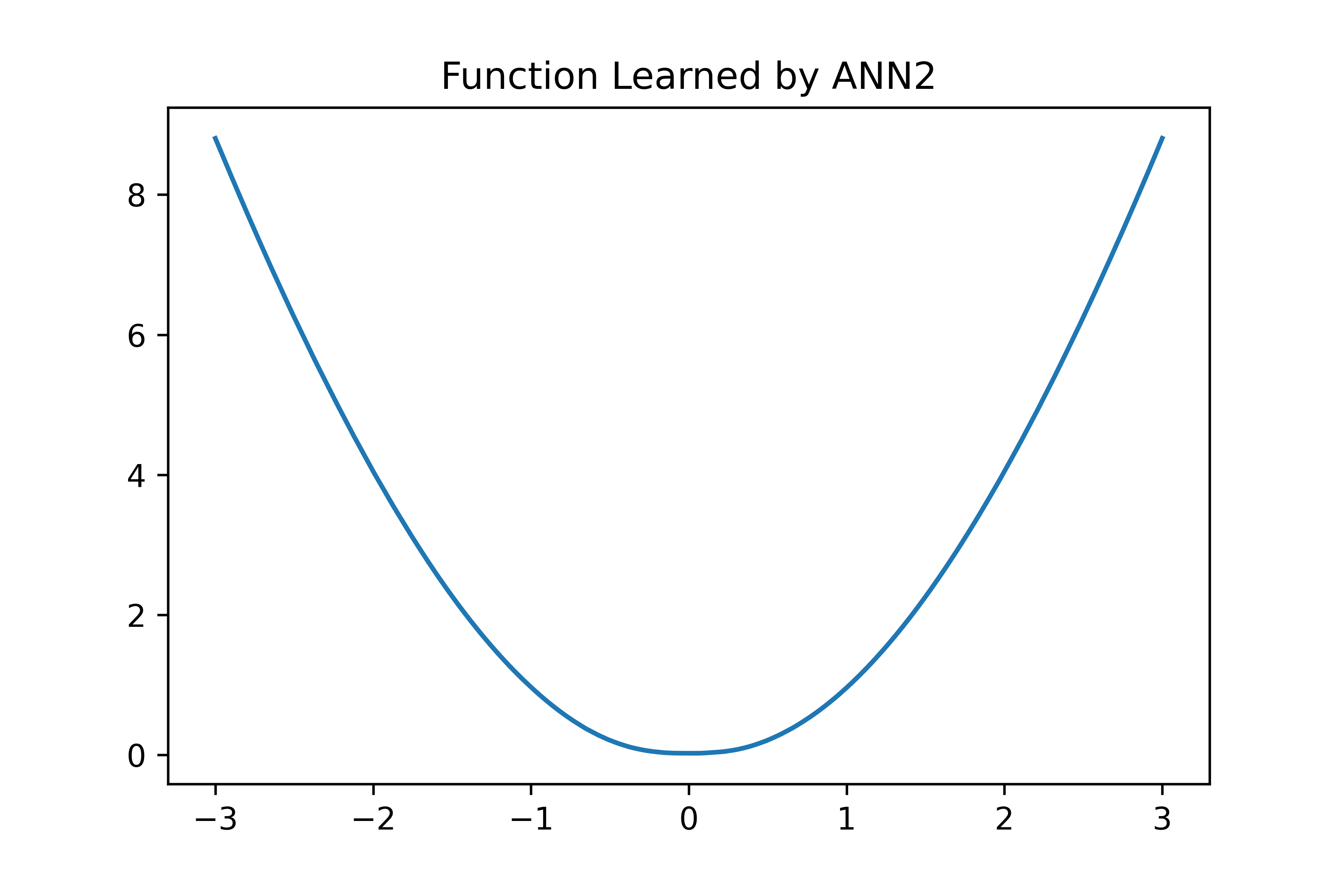}
            }
            \subfloat[$f_3(x_1,x_2;\theta_3)$]{
                \includegraphics[width=0.31\linewidth]{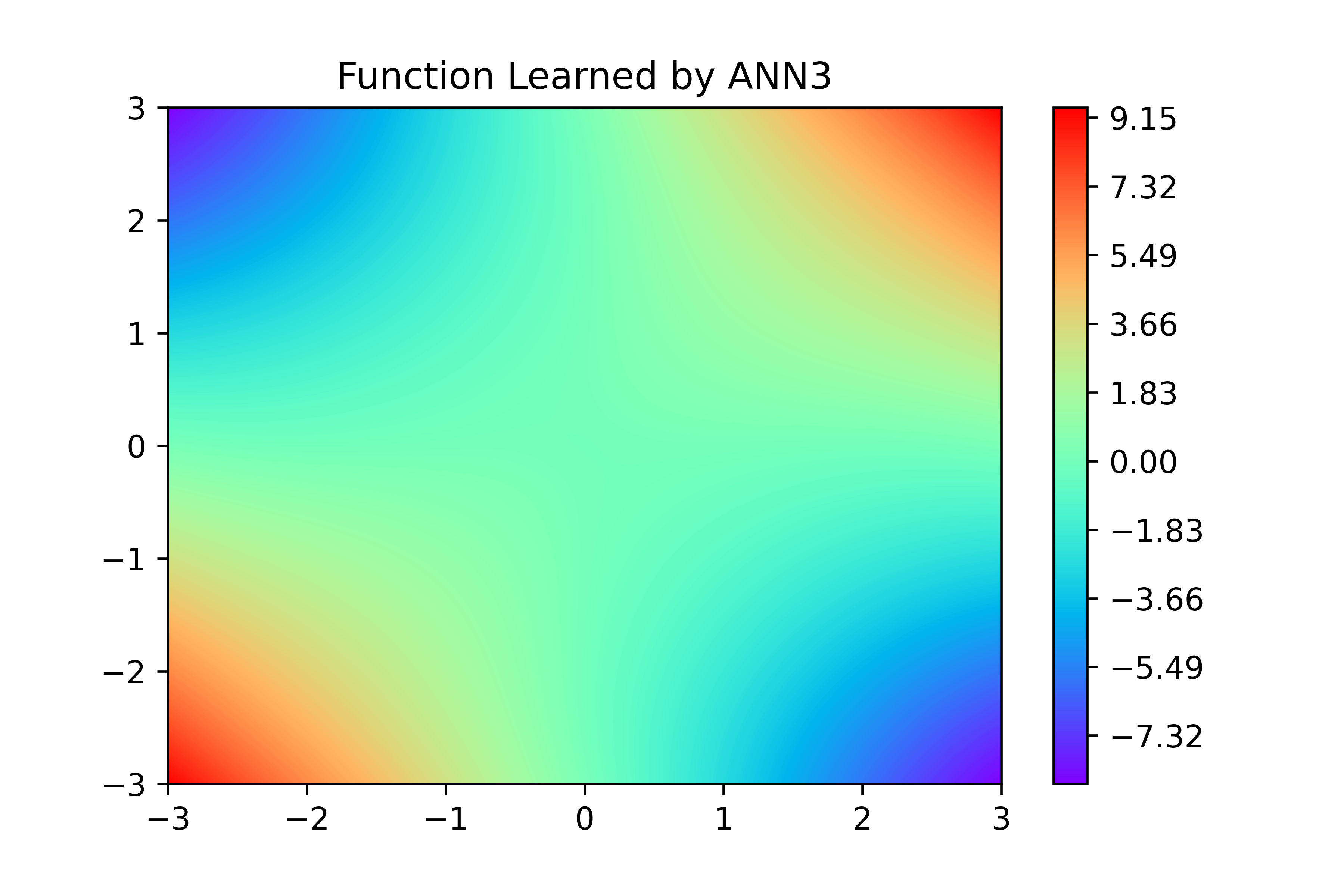}
            }
            \setcounter{subfigure}{2}
            \label{fig3:b}
        }
    \label{Fig3_a}
    \end{minipage}
    \begin{minipage}{0.38\textwidth}
        \subfloat[PAL]{
            \includegraphics[width=1.0\linewidth]{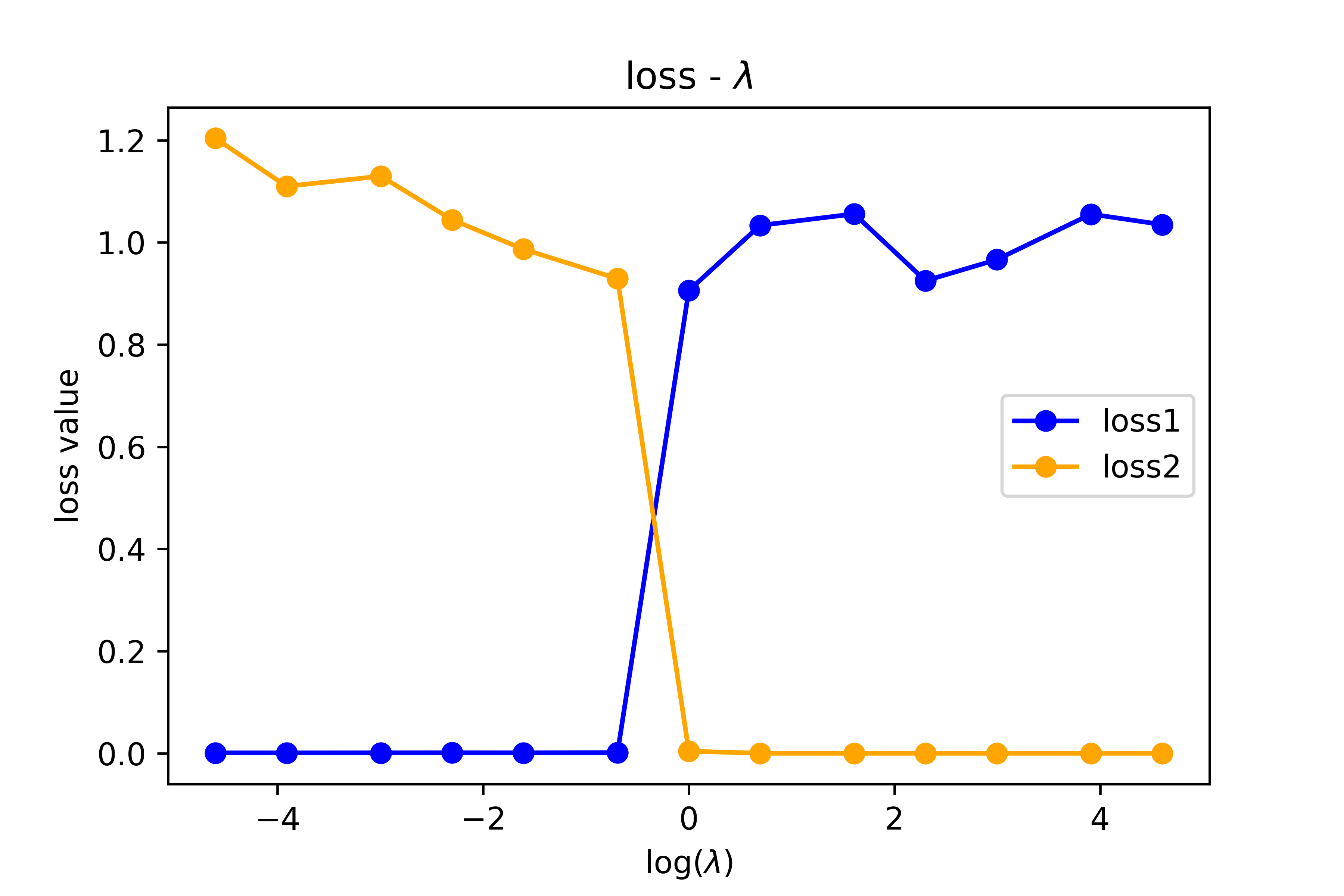}
            \label{fig3:c}
        }\\
        \subfloat[PIL]{
            \includegraphics[width=1.0\linewidth]{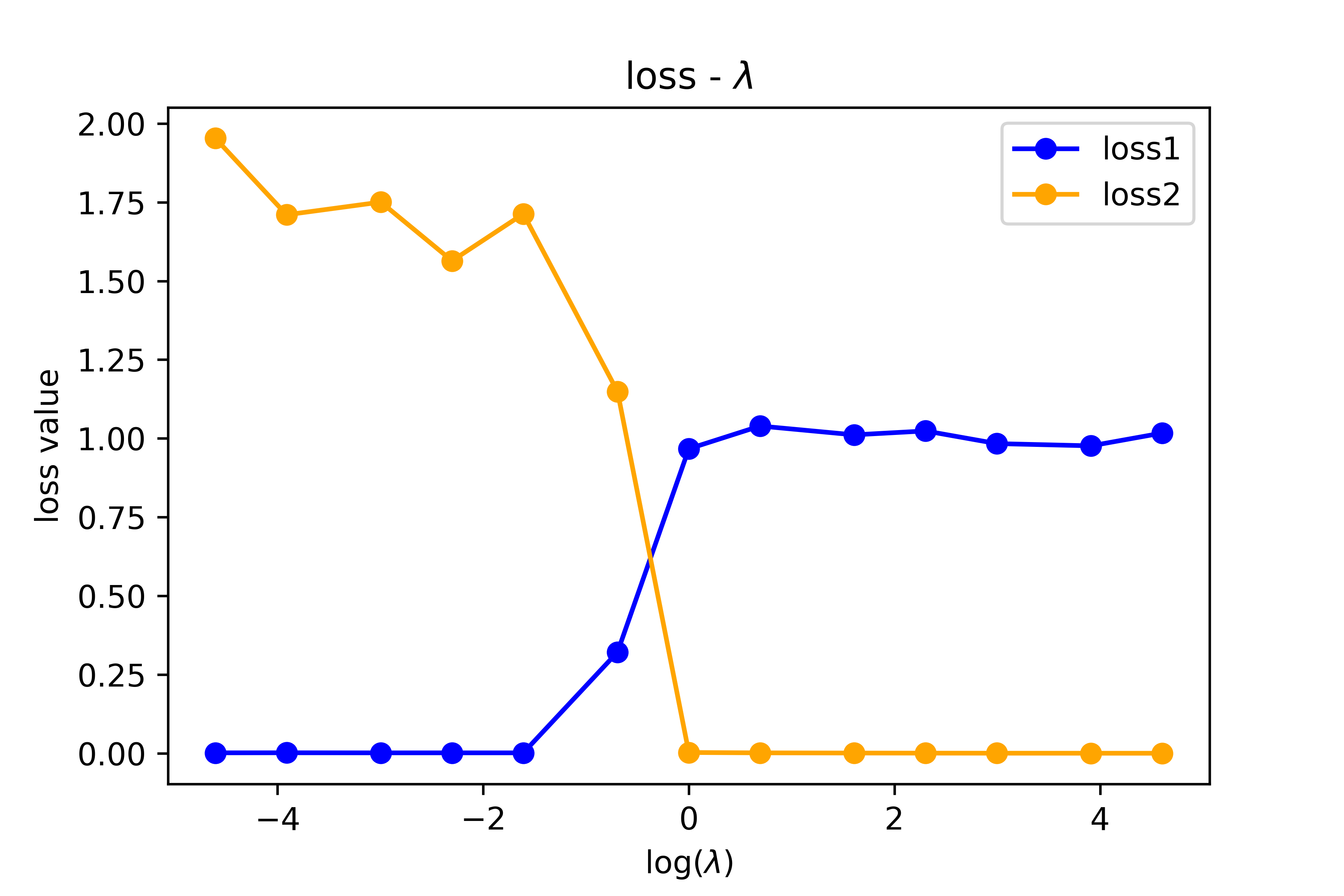}
            \label{fig3:d}
        }
    \end{minipage}
    \caption{(a)(b) Additive separability learned by PAL; (c)(d) $\lambda$-loss relations in PAL and PIL.}
    \label{fig3}
\end{figure}

We employ PAL To obtain the decomposition numerically first. we train three neural networks, $f_1(x_1;\theta_1)$, $f_2(x_2;\theta_2)$, and $f_{12}(x_1, x_2;\theta_{12})$, with $\lambda=0.2$ for 200 epochs. We use the ADAM optimizer and annealed learning rate schedule i.e., $\{10^{-3},10^{-4},10^{-5},10^{-6}\}$ each learning rate for 50 epochs. We show three neural networks succeed in learning the ground truth decomposition in Figure \ref{fig3}\subref{fig3:a}-\subref{fig3:b}. 
After PAL training, we then apply AI Feynman to explain the outputs i.e., symbolic expression of these three neural networks: $f_1=x_1^2-0.02$, $f_2=x_2^2-0.01$, and $f_{12}=x_1x_2+0.03$.

{\bf Phase transition} A recent work~\cite{nnphd} theoretically proves that $L_1,L_2\sim \lambda$ has a phase transition behavior at $\lambda=1$~\footnote{Loss functions $L_1$, $L_2$ should be defined as norms to produce the sharp phase transition behavior, e.g., mean-absolute error (MAE) or square root of mean-squared error (MSE). By contrast, MSE does not produce any sharp phase transition}. $\lambda>1$ is the undesireable phase, so in principle one can simply choose any $\lambda<1$ to obtain the correct result. Their numerical results suggest that $\lambda\in[0.02,0.5]$ produce very accurate and robust results. We verify these observations in the current example.

We sweep the $\lambda$ region and obtained final losses ($L_1$ and $L_2$) as a function of $\lambda$ in PAL and PIL, by testing $\lambda=\left\{0.01,0.02,0.05,0.1,0.2,0.5,1,2,5,10,20,50,100\right\}$. We show the results in Figure \ref{fig3}\subref{fig3:c}-\subref{fig3:d}. For PAL, there is a sharp phase transition at $\lambda=1$ and two losses are quite robust for any $\lambda<1$. Intriguingly, the sharp phase transition for PIL has never been reported let alone studied before (to the best of knowledge), which will be interesting future directions.

{\bf Rotational invariance}: We use PAL to decompose the function $f_0(x_1,x_2)=\frac{1}{2}(x_1^2+x_2^2)+ax_1$ into the rotational-invariant part $\frac{1}{2}(x_1^2+x_2^2)$ and violation part $ax_1$ $(a=0.32)$. In the experiment, we take $\lambda=0.2$ and jointly train $f_1(R=\sqrt{x_1^2+x_2^2};\theta_1)$ and $f_2(x_1,x_2;\theta_2)$ for 200 epochs. We use the ADAM optimizer and annealed learning rate schedule i.e., $\{10^{-3},10^{-4},10^{-5},10^{-6}\}$ each learning rate for 50 epochs.

We apply AI Feynman to explain $f_1$ and $f_2$ parameterized by neural networks: AI Feynman discovers that $f_1=0.5(x_1^2+x_2^2)$ and $f_2=0.31998x_1$.

{\bf Positive Definiteness}: Given a function $f_0(x)={\rm sin}({\rm sin}(x))+0$, we train the the first neural network in a nested form $f_1(f_1(x;\theta_1);\theta_1)$ for the positivity part and the second neural network for violation part. We jointly train two networks for 2000 epochs with the ADAM optimizer. We employ annealed learning rate schedule i.e., $\{10^{-3},10^{-4},10^{-5},10^{-6}\}$ each learning rate for 500 epochs. AI Feynman discovers the symbolic expression for two networks: $f_1=-\sin(x)+0.004$ and $f_2=0$.

{\bf PIL for additive separability and rotational invariance}: To compare PAL and PIL, we also use PIL to learn the additive separability and rotational invariance under the same setting. The results of symbolic regression are $x_1^2+x_2^2+x_1x_2$ (additive separability) and $0.5(x_1^2+x_2^2)+0.32x_1$ (rotational invariance). Although PIL can obtain correctly the whole symbolic expression, it does not support decompositions into the property part and the violation part. For the positivity example, PIL is even not applicable since positivity is non-discriminative.

\section{N-body Dynamics Details}\label{app:n-body}

{\bf Physical Model}

We consider a 2D $N$-particle system described by their positions $\mat{x}_i$ and velocities $\mat{v}_i$ $(i=1,\cdots,N)$. They all have unit mass $m=1$ and have pairwise forces i.e., particle $i$ exerts to particle $j$ a force $\mat{f}_{ij}=f(r_{ij})\frac{\mat{r}_{ij}}{r_{ij}}$ where $\mat{r}_{ij}\equiv \mat{x}_i-\mat{x}_j$, $r_{ij}\equiv |\mat{r}_{ij}|_2$ and $f(\cdot)$ is the same for all pairs.

The dynamical equations for the $j$-th particle is:

\begin{equation}
    \begin{aligned}
    \frac{d}{dt}
    \begin{pmatrix}
    \mat{x}_j \\
    \mat{v}_j \\
    \end{pmatrix}
    =
    \begin{pmatrix}
    \mat{v}_j \\
    \sum_{i\neq j}\mat{f}_{ij}
    \end{pmatrix}
    \end{aligned}
\end{equation}

Concatenating all the particle states together defines $\mat{z}=[\mat{x}_1,\cdots,\mat{x}_N,\mat{v}_1,\cdots,\mat{v}_N]$

\begin{equation}\label{eq:N-body-F}
    \begin{aligned}
    \frac{d}{dt}\mat{z}\equiv
    \frac{d}{dt}
    \begin{pmatrix}
    \mat{x}_1 \\
    \cdots \\
    \mat{x}_N \\
    \mat{v}_1 \\
    \cdots \\
    \mat{v}_N \\
    \end{pmatrix}
    =
    \begin{pmatrix}
    \mat{v}_1 \\
    \cdots \\
    \mat{v}_N \\
    \sum_{i\neq 1}\mat{f}_{i1}\\
    \cdots \\
    \sum_{i\neq N}\mat{f}_{iN}\\
    \end{pmatrix}
    \equiv \mat{F}(\mat{z};f)
    \end{aligned}
\end{equation}

Discretizing Eq.~(\ref{eq:N-body-F}) in time gives 

\begin{equation}\label{eq:N-body-discrete}
    \mat{z}^{l+1} = \mat{z}^l + \mat{F}(\mat{z};f)\Delta t \quad (\mat{z}^l\equiv \mat{z}(t=l\Delta t))
\end{equation}

In our simulation, the initial position and initial velocity of each particle are randomly sampled from a standard Gaussian distribution $\mat{x}_i \sim \mathcal{N}([0,0]^T,\,\sigma^{2}\mat{I}_{2\times 2})$ where $\sigma=1$. The force between each pair of two bodies is set as $f(r)=r^2$. We use Newton forward scheme as in Eq.~(\ref{eq:N-body-discrete}) to simulate the 5-body system for $50$ steps with step size $\Delta t = 0.02$. The initial positions for the 5 bodies are $(1.62,-0.61)$, $(-0.53,-1.07)$, $(0.87,-2.30)$, $(1.74,-0.76)$, $(0.32,-0.25)$. 
The initial velocities for the 5 bodies are $(2.92,-4.12)$, $(-0.64,-0.77)$, $(2.27,-2.20)$, $(-0.34,-1.76)$, $(0.08,1.17)$. 

{\bf Neural network architecture}

Note that our goal is to infer the interaction $f(\cdot)$ based on solely the initial $\mat{z}(t=0)$ and final $\mat{z}(t=1)$. Our network is effectively a residual network with $50$ blocks -- it simply implements the computations as in  Eq.~(\ref{eq:N-body-discrete}), with only $f$ parameterized by neural networks. To impose $f$ to be time-independent, PAL and PIL employ different strategies. 

\textbf{PAL}: explicit decomposition $f(r,t)=f_1(r;\theta_1)+f_2(r,t;\theta_2)$ and penalize $|f_2(r,t;\theta_2)|$. Both $f_1$ and $f_2$ are 2-hidden-layer fully-connected networks (tanh activation) with each layer containing $(1/2,200,200,1)$ neurons. we take $\lambda_1=1$, $\lambda_2=0.25$, and $\lambda_3=1$, where $\lambda_1$ is the weight for the final position prediction loss, $\lambda_2$ is the weight for the final velocity prediction loss, and $\lambda_3$ penalty coefficient for $f_2(r,t;\theta_2)$. We train two neural networks $f_1(r;\theta_1)$ and $f_2(r,t;\theta_2)$, each repeated $100$ times and stacked vertically to obtain a 100-layer Resnet. Both $f_1$ and $f_2$ are two-hidden-layer fully-connected networks (Activation function: tanh) with each layer containing $(1/2,200,200,1)$ neurons. We train two networks jointly for $2000$ epochs with the Adam optimizer with the learning rate $0.001$.

\textbf{PIL}: no explicit decomposition but penalize $|f(r,t_1)-f(r,t_2)|$ for $t_1\neq t_2$. Each block is a 2-hidden-layer fully-connected network (tanh activation) with each layer containing $(1/2,200,200,1)$ neurons. For the loss function, we take $\lambda_1=1$, $\lambda_2=0.25$, and $\lambda_3=0.1$, where $\lambda_1$ is the weight for the final position prediction loss, $\lambda_2$ is the final velocity prediction loss, and $\lambda_3$ is the time-independence penalty loss. The time-independence loss is simply implemented as the MSE between the outputs of two neighboring blocks of the Resnet for the same input. We train the Resnet for $2000$ epochs with an Adam optimizer with the learning rate $0.0001$.



\end{appendix}

\end{document}